\documentclass[a4paper]{article}
\usepackage{spconf,amssymb,amsmath,graphicx,epsfig}
\ninept	

\title{Optimal Size, Freshness and Time-frame for Voice Search Vocabulary}

\makeatletter
\def\name#1{\gdef\@name{#1\\}}
\makeatother
\name{{\em Maryam Kamvar,  Ciprian Chelba}}

\address{Google, Inc., 1600 Amphitheatre Parkway, Mountain View, CA 94043, USA \\
  {\small \tt \{mkamvar,ciprianchelba\}@google.com}}

\begin{document}
\maketitle

\begin{abstract}
In this paper, we investigate how to optimize the vocabulary for a
voice search language model. The metric we optimize over is the
out-of-vocabulary (OoV) rate since it is a strong indicator of user
experience. In a departure from the usual way of measuring OoV rates,
web search logs allow us to compute the per-session OoV rate and thus
estimate the percentage of users that experience a given OoV
rate. Under very conservative text normalization, we find  that a
voice search vocabulary consisting of 2 to 2.5M words extracted from 1
week of search query data will result in an aggregate OoV rate of 0.01;
at that size, the same OoV rate will also be experienced by 90\% of
users. The number of words included in the vocabulary is a stable
indicator of the OoV rate. Altering the \emph{freshness} of the
vocabulary or the duration of the time window over which the
training data is gathered does not significantly change the OoV
rate. Surprisingly, a significantly larger vocabulary (approx.\ 10
million words) is required to guarantee OoV rates below 0.01 (1\%) for
95\% of the users.
\end{abstract}
\vspace{0.25cm}

\section{Introduction}
The OoV rate is one indication of user experience in
voice search, and automatic speech recognition (ASR) in general; the
higher the OoV rate, the more likely the user is to have a poor
experience. Clearly, each OoV word will result in at least one error
at the word level\footnote{The approximate rule of thumb is 1.5
  errors for every OoV word, so an OoV rate of 1\% would lead to about
  1.5\% absolute loss in word error rate (WER).}, and in exactly one
error at the whole query/sentencelevel. In ASR practice, OoV rates below
0.01 (1\%) are deemed acceptable since typical WER values are well above
10\%.

As shown in~\cite{chelba:slt2010}, a typical vocabulary for a US
English voice search language model (LM) is trained on the US English query
stream, contains about one million words, and achieves
out-of-vocabulary (OoV) rate of 0.57\% on unseen text query data,
after query normalization.

In a departure from typical vocabulary estimation
methodology, \cite{jelinek1990self,venkataraman2003techniques}, the web
search query stream not only provides us with training data for the
LM, but also with session level information based on 24-hour
cookies. Assuming that each cookie corresponds to the experience of a
web search user over exactly one day, we can compute per-one-day-user
OoV rates, and directly corelate them with the voice search LM
vocabulary size.

Since the vocabulary estimation algorithms are extremely simple, the
paper is purely experimental. Our methodology is as follows:
\begin{itemize}\addtolength{\itemsep}{-0.5\baselineskip}
\item select as training data $\mathcal{T}$ a set of queries arriving at the
  \verb+google.com+ front-end during time period $T$;
\item text normalize the training data, see Section~\ref{tn};
\item estimate a vocabulary $\mathcal{V}$ by thresholding the 1-gram
  count of words in $\mathcal{T}$ such that it exceeds $C$,
  $\mathcal{V}(T, C)$;
\item select as test data $\mathcal{E}$ a set of queries arriving at the
  \verb+google.com+ front-end during time period $E$; $E$ is a single
  day that occurs after $T$, and the data is subjected to the exact
  same text normalization used in training;
\item we evaluate both \emph{aggregate} and \emph{per-cookie} OoV
  rates, and report the aggregate OoV rate across all words in
  $\mathcal{E}$, as well as the percentage of cookies in $\mathcal{E}$
  that experience an OoV rate that is less or equal than 0.01 (1\%).
\end{itemize}
We aim to answer the following questions:
\begin{itemize}\addtolength{\itemsep}{-0.5\baselineskip}
\item how does the vocabulary size (controlled by the threshold $C$)
  impact both \emph{aggregate} and \emph{per-cookie} OoV rates?
\item how does the vocabulary freshness (gap between $T$ and $E$) impact the OoV rate?
\item how does the time-frame (duration of $T$) of the training data
  $\mathcal{T}$ used to estimate the vocabulary $\mathcal{V}(T, C)$ impact the
  OoV rate?
\end{itemize}

\section{A note on query normalization}
\label{tn}
We build the vocabulary by considering all US English queries logged during
$T$. We break each query up into words, and discard words that have
non-alphabetic characters. We perform the same normalization for the
test set. So for example if the queries in $\mathcal{T}$ were: 
\verb+gawker.com, pizza san francisco,+\\\verb+baby food, 4chan status+ the
resulting vocabulary would be \verb+pizza, san, francisco, baby, food,+\\\verb+status+. 
The query \verb+gawker.com+ and the word \verb+4chan+ would not be
included in the vocabulary because they contain non-alphabetic
characters.

We note that the above query normalization is extremely conservative
in the sense that it discards a lot of problematic cases, and keeps
the vocabulary sizes and OoV rates smaller than what would be required
for building a vocabulary and language model that would actually be
used for voice search query transcription. As a result, the vocabulary
sizes that we report to achieve certain OoV values are very likely
just lower bounds on the actual vocabulary sizes needed, were correct
text normalization (see~\cite{chelba:slt2010} for an example text
normalization pipeline) to be performed.

\section{Experiments}
\label{exp}

The various vocabularies used in our experiment are created from
queries issued during a one-week to one-month period starting on
10/04/2011. The vocabulary is comprised of the words that were
repeated $C$ or more times in $\mathcal{T}$. We chose seven values for
$C$: 960, 480, 240, 120, 60, 30 and 15. As $C$ decreases, the vocabulary
size increases; to preserve user privacy we do not use $C$ values lower
than 15. For each training set $\mathcal{T}$ discussed in this paper,
we will create seven different vocabularies based on these thresholds.

Each test set $\mathcal{E}$ is comprised of queries associated with a set of
over 10 million cookies during a one-day period. We associate test queries
by cookie-id in order to compute user-based (per-cookie) OoV rate.

All of our data is strictly anonymous; the queries bear no user-identifying
information. The only query data saved after training are the
vocabularies. The evaluation on test data is done by counting on
streamed filtered query logs, without saving any data.

\begin{figure*}[p]
  \begin{center}
    \epsfig{figure=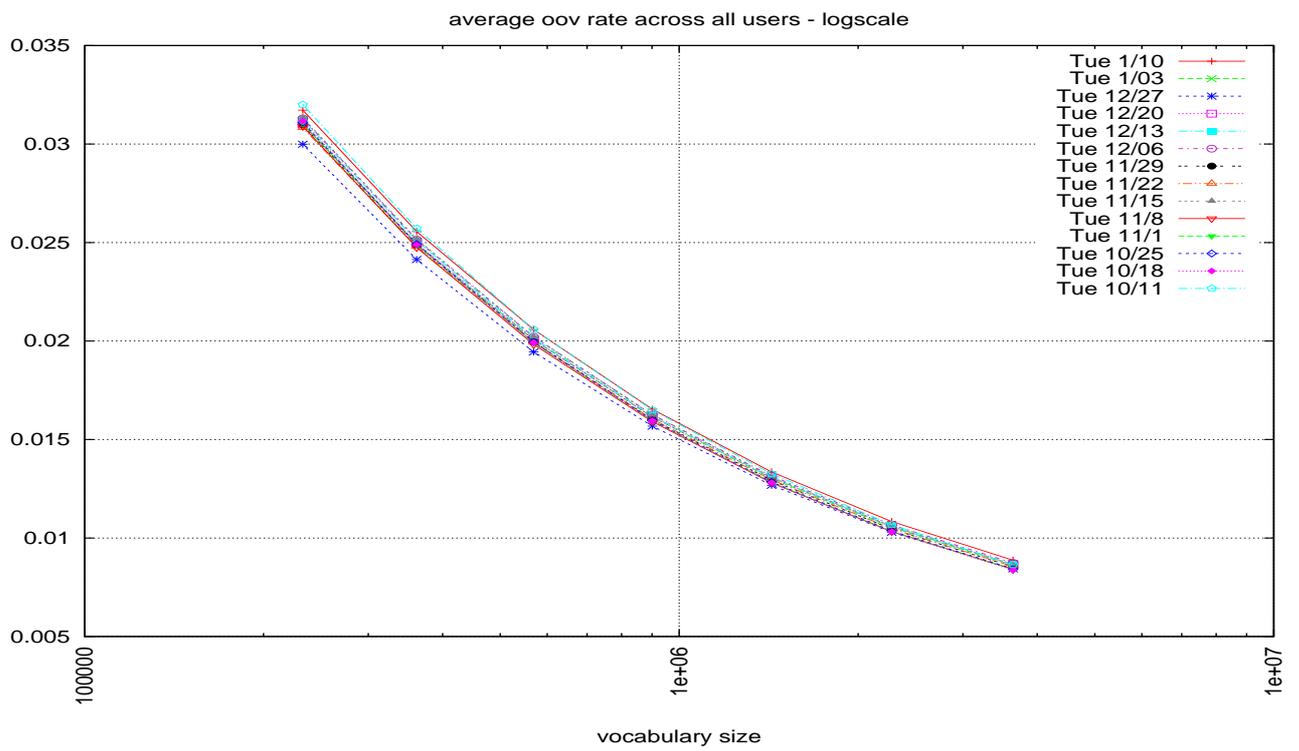,width=2.1\columnwidth,height=10cm}
    \caption{{\it Aggregate OoV Rate as a Function of Vocabulary Size
        (log-scale), Evaluated on a Range of Test Sets Collected every
        Tuesday between 2011/10/11-2012/01/03.}}  
    \label{avg_oov_tue}
  \end{center}
\end{figure*}

\begin{figure*}[p]
  \begin{center}
    \epsfig{figure=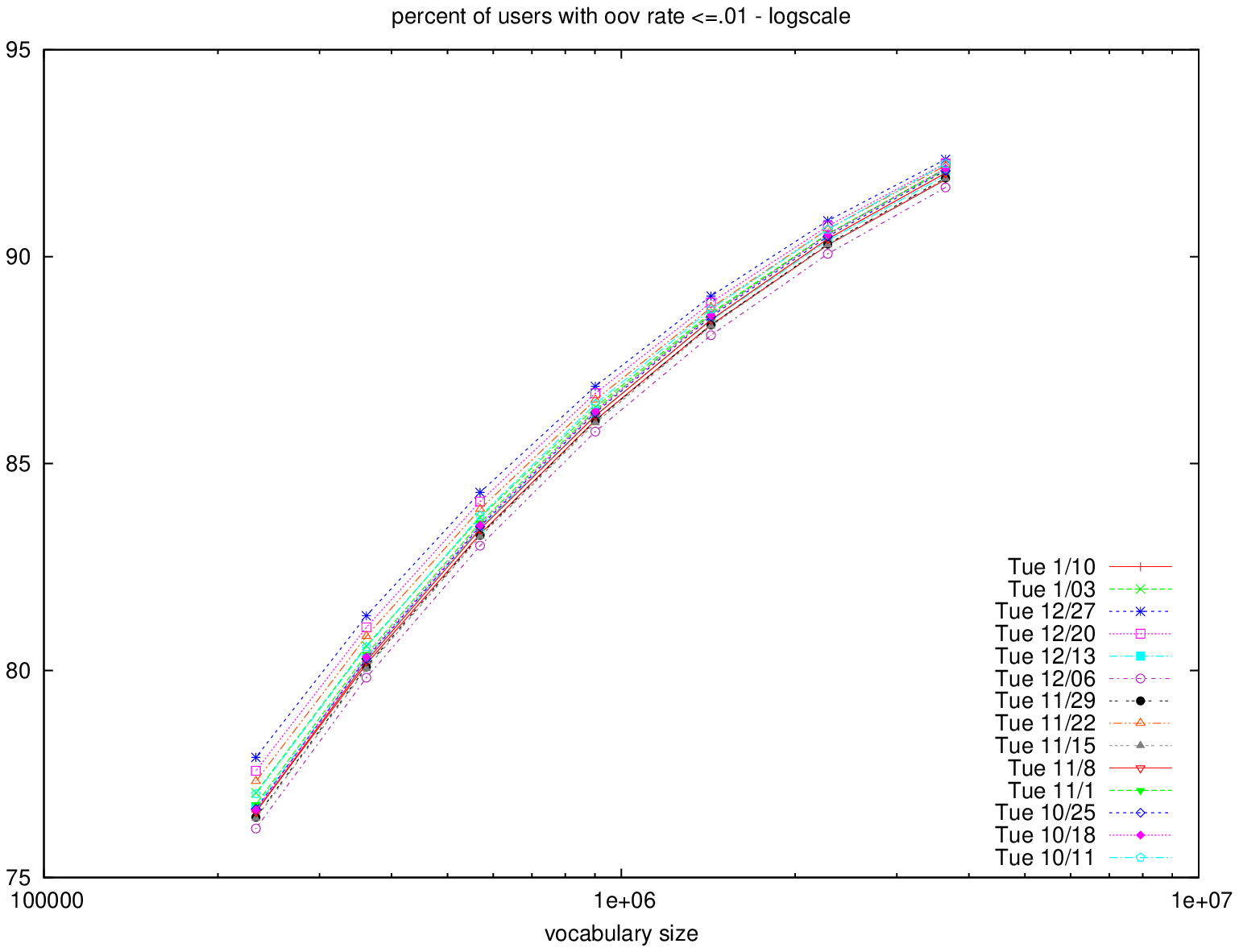,width=2.1\columnwidth,height=10cm}
    \caption{{\it Percentage of Cookies/Users with OoV Rate less than
        0.01 (1\%) as a Function of Vocabulary Size (log-scale), Evaluated on
        Test Sets Collected every Tuesday between
        2011/10/11-2012/01/03.}}  
    \label{cookie_oov_tue}
  \end{center}
\end{figure*}

\subsection{Vocabulary Size}
\label{exp:voc_size}

To understand the impact of vocabulary size on OoV rate, we created
several vocabularies from the queries issued in the week
$T=10/4/2011-10/10/2011$. The size of the various vocabularies as a
function of the count threshold is presented in Table~\ref{vocab};
Fig.~\ref{avg_oov_tue} shows the relationship
between the logarithm of the size of the vocabulary and the aggregate
OoV rate---a log-log plot of the same data points would reveal a
"quasi-linear" dependency.
\begin{table}[t,h]
\caption{\label{vocab} {\it Vocabulary size as a function of count threshold.}}
\vspace{2mm}
\centerline{
\begin{tabular}{|r|r|}
\hline
threshold & vocabulary size\\\hline
15	&	3,643,583\\
30	&	2,277,696\\
60	&	1,429,888\\
120	&	901,213\\
240	&	569,330\\
480	&	361,776\\
960	&	232,808\\\hline
\end{tabular}}
\end{table}
We have also measured the percentage of cookies/users for a given OoV
rate (0.01, or 1\%), and the results are shown in
Fig.~\ref{cookie_oov_tue}. At a vocabulary size of 2.25 million words
($C=30$, aggregate OoV=0.01), over 90\% of users will experience an OoV
rate of 0.01.
\subsection{Vocabulary Freshness}

To understand the impact of the vocabulary freshness on the OoV rate,
we take the seven vocabularies described above ($T = 10/4/2011 -
10/10/2011$ and $C = 960, 480, 240, 120, 60, 30, 15$) and investigate
the OoV rate change as the lag between the training data $T$ and the
test data $E$ increases: we used the 14 consecutive Tuesdays between
$2010/10/11-2011/01/20$ as test data. We chose to keep the day of week
consistent (a Tuesday) across this set of $E$ dates in order to
mitigate any confounding factors with regard to day-of-week.

We found that within a 14-week time span, as the \emph{freshness} of
the vocabulary decreases, there is no consistent increase in the
aggregate OoV rate (Fig.~\ref{avg_oov_tue}) nor any significant
decrease in the percentage of users who experience less than 0.01 (1\%) OoV
rate (Fig.~\ref{cookie_oov_tue}).

\subsection{Vocabulary Time Frame}

\begin{figure*}[p]
  \begin{center}
    \epsfig{figure=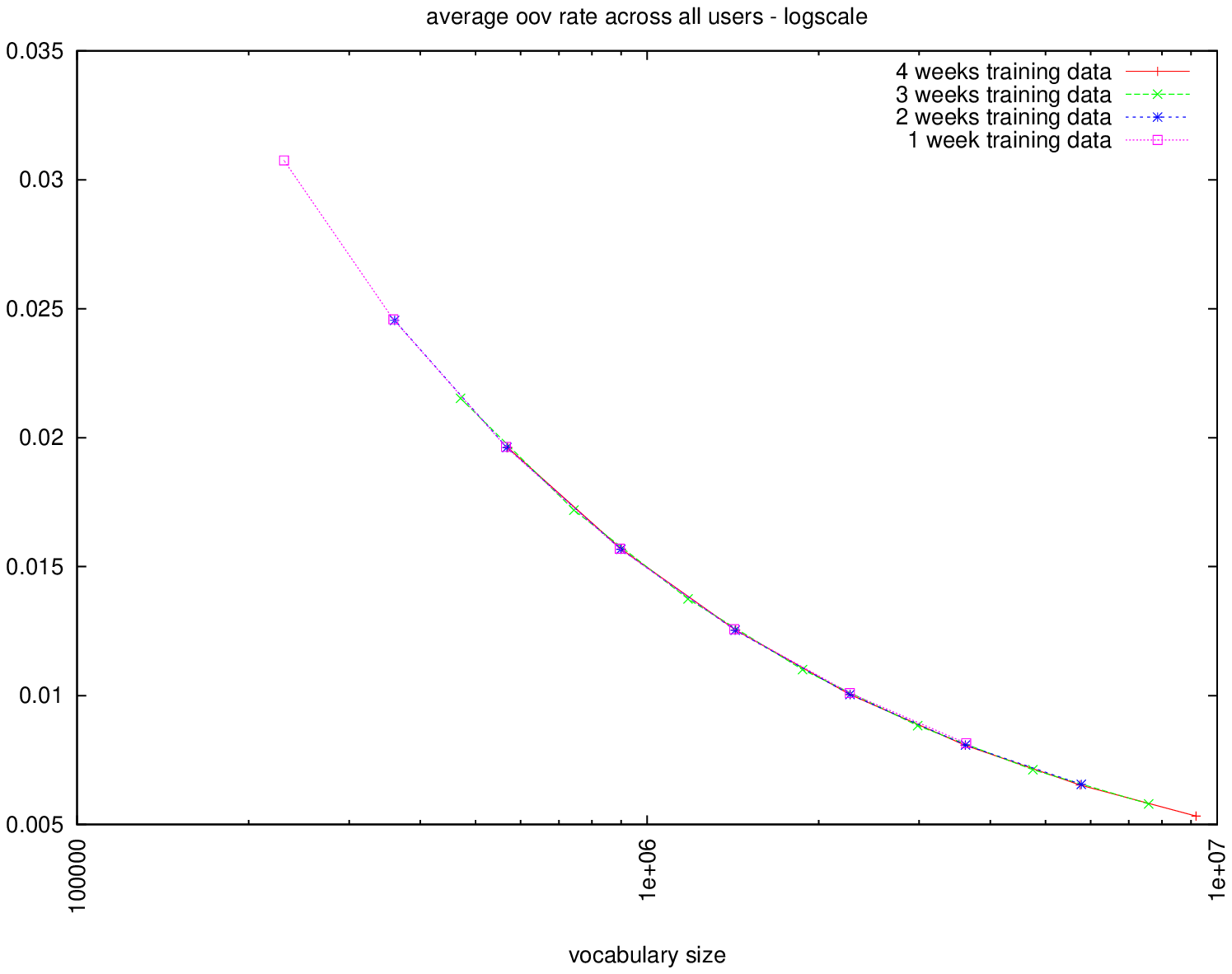,width=2.1\columnwidth,height=10cm}
    \caption{{\it Aggregate OoV rate on 11/1/2011 over Vocabularies Built from Increasingly Large Training Sets.}}  
    \label{avg_var_t}
  \end{center}
\end{figure*}

\begin{figure*}[p]
  \begin{center}
    \epsfig{figure=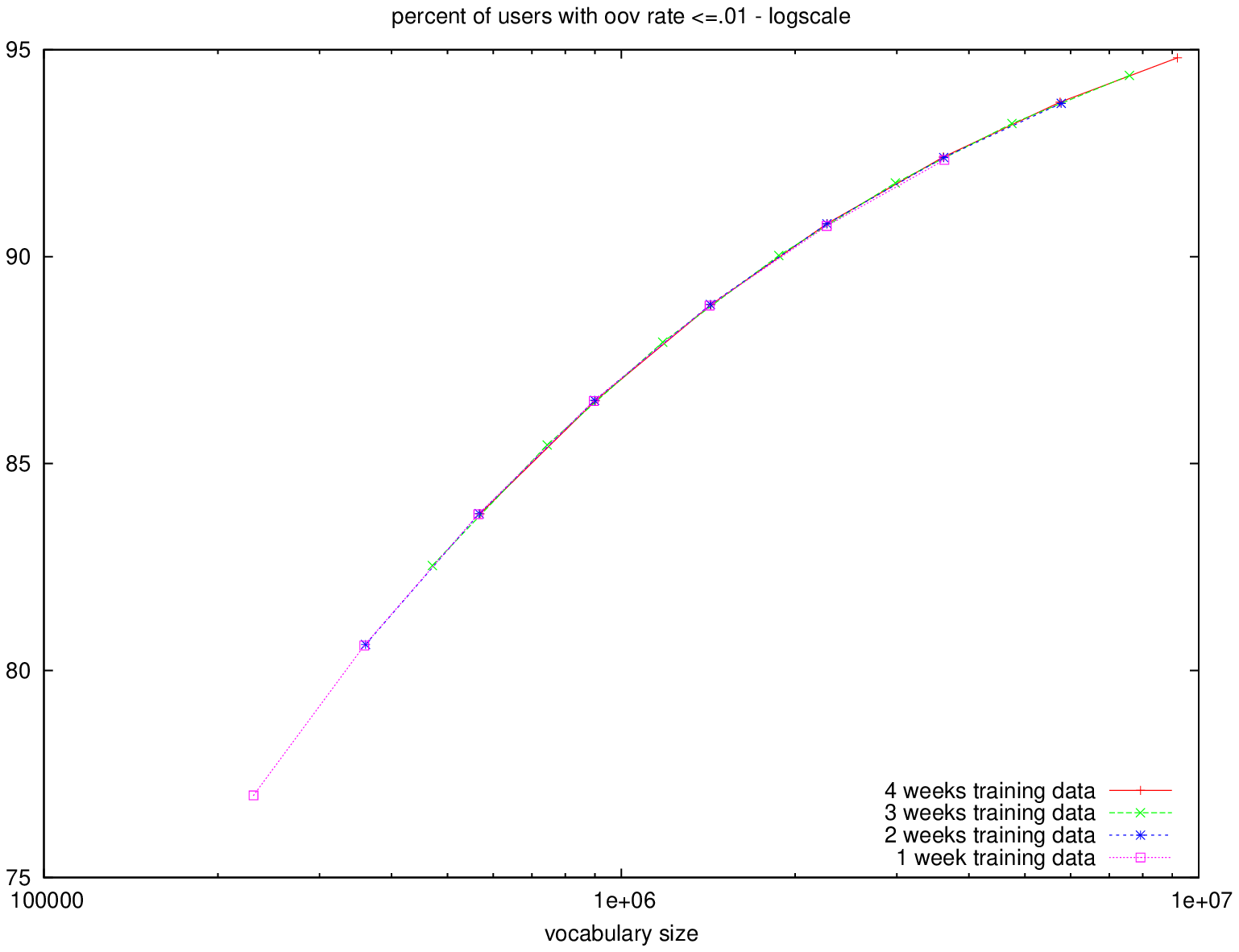,width=2.1\columnwidth,height=10cm}
    \caption{{\it Percentage of Cookies/Users with OoV Rate less than
        0.01 (1\%) on 11/1/2011 over Vocabularies Built from Increasingly Large Training Sets.}}  
    \label{cookie_var_t}
  \end{center}
\end{figure*}

To understand how the duration of $T$ (the time window over which the
vocabulary is estimated) impacts OoV rate, we created vocabularies
over the following time windows:
\begin{itemize}\addtolength{\itemsep}{-0.5\baselineskip}
\item 1 week period between $10/25/2011 - 10/31/2011$
\item 2 week period between $10/18/2011 - 10/31/2011$
\item 3 week period between $10/11/2011 - 10/31/2011$
\item 4 week period between $10/04/2011 - 10/31/2011$
\end{itemize}
We again created seven threshold based vocabularies for each $T$. We
evaluate the aggregate OoV rate on the date $E = 11/1/2011$, see
Fig.~\ref{avg_var_t}, as well as the percentage of users with a
per-cookie OoV rate below 0.01 (1\%), see Fig.~\ref{cookie_var_t}.
We see that the shape of the graph is fairly consistent across $T$ time
windows, and a week of training data is as good as a month.

More interestingly, Fig.~\ref{cookie_var_t} shows that aiming at an
operating point where 95\% the percentage of users experience OoV rates below
0.01 (1\%) requires significantly larger vocabularies, approx.\ 10
million words.

\section{Conclusions}
To guarantee out-of-vocabulary rates below 0.01 (1\%) we find that we need a
vocabulary of 2-2.5 million words (even with the conservative text
normalization described in Section~\ref{tn}). That vocabulary size
guarantees OoV rates below 0.01 (1\%) for 90\% of the users. 

Somewhat surprisingly, our experimental data shows that a significantly larger
vocabulary (approx.\ 10 million words) seems to be  required to
guarantee a 0.01 (1\%) OoV rate for 95\% of the users.

Studies on the \verb+www+ pages side~\cite{brants2012} show that after
just a few million words, vocabulary growth is close to a straight
line in the logarithmic scale; the vocabulary grows by about 69\% each
time the size of the text is doubled even when using 1 trillion words
of training data. Since queries are used for finding such pages, the
growth in query stream vocabulary size is easier to understand.

We also find that one week of data is as good as one month for
estimating the vocabulary, and that there is very little drift in OoV rate
as the test data (one day) shifts during the three months following
the training data used for estimating the vocabulary.

\section{Acknowledgements}
We would like to thank Thorsten Brants for useful discussions.

\bibliographystyle{IEEEbib}
\bibliography{../../../../../../mainbibfile}
\end{document}